\documentclass[runningheads]{llncs}
\usepackage[T1]{fontenc}
\usepackage{graphicx,verbatim}
\usepackage{amsmath}
\usepackage{amsfonts}
\usepackage{dsfont}
\usepackage{graphicx}
\usepackage{tikz}
\usepackage{multirow}
\usepackage{array}
\usepackage{tabularx,booktabs}
\usepackage{subcaption}
\usepackage{dirtytalk}
\usepackage{bm}
\usepackage{dsfont}
\usepackage{float}
\usetikzlibrary{arrows.meta}
\usepackage{color}
\usepackage{hyperref}
\usepackage{rotating}
\usepackage{mathrsfs}
\usepackage[misc]{ifsym}

\definecolor{cgbla}{rgb}{0.5, 0, 0.5}
\definecolor{cliver}{rgb}{0.2, 0.8, 0.85}
\definecolor{cgrasper}{rgb}{0.2, 0.85, 0.3}

\begin{document}
\newcommand{\repeatthanks}{\textsuperscript{\thefootnote}}

\title{When Tracking Fails: Analyzing Failure Modes of SAM2 for Point-Based Tracking\\in Surgical Videos}
\titlerunning{Failure Modes of SAM2 for Point-Based Tracking in Surgical Videos}

\author{
Woowon Jang\inst{1}\thanks{Equal contribution.} \and
Jiwon Im\inst{1}\repeatthanks \and
Juseung Choi\inst{1}\repeatthanks \and
Niki Rashidian\inst{2,3} \and\\
Wesley De Neve\inst{1,4} \and
Utku Ozbulak\inst{1,4}(\Letter)
}

\authorrunning{Jang et al.}

\institute{
Center for Biosystems and Biotech Data Science, Ghent University Global Campus, Incheon, Republic of Korea \and
Department of Human Structure and Repair, Ghent University, Ghent, Belgium \and
Department of HPB Surgery \& Liver Transplantation, Ghent University Hospital, Ghent, Belgium \and
IDLab, Department of Electronics and Information Systems, Ghent University, Ghent, Belgium\\
(\Letter) \email{utku.ozbulak@ghent.ac.kr}
}

\maketitle
\begin{abstract}
\let\thefootnote\relax\footnotetext{Accepted for publication in the 28th International Conference on Medical Image Computing and Computer Assisted Intervention (MICCAI) Workshop on Collaborative Intelligence and Autonomy in Image-guided Surgery (COLAS), 2025.}
Video object segmentation (VOS) models such as SAM2 offer promising zero-shot tracking capabilities for surgical videos using minimal user input. Among the available input types, point-based tracking offers an efficient and low-cost alternative, yet its reliability and failure cases in complex surgical environments are not well understood. In this work, we systematically analyze the failure modes of point-based tracking in laparoscopic cholecystectomy videos. Focusing on three surgical targets, the gallbladder, grasper, and L-hook electrocautery, we compare the performance of point-based tracking with segmentation mask initialization. Our results show that point-based tracking is competitive for surgical tools but consistently underperforms for anatomical targets, where tissue similarity and ambiguous boundaries lead to failure. Through qualitative analysis, we reveal key factors influencing tracking outcomes and provide several actionable recommendations for selecting and placing tracking points to improve performance in surgical video analysis.
\end{abstract}

\section{Introduction}

Artificial intelligence (AI)-based technologies are increasingly being integrated into surgical workflows to enhance intraoperative decision-making, post-operative analysis, and practitioner training~\cite{10.1001/jamasurg.2019.4917,mousavi2024reference}. Among these, AI-based surgical video understanding has gained particular attention due to its potential to automate annotation, identify anatomical structures, and provide context-aware guidance during surgical procedures~\cite{cancers16101870}. The ability to interpret visual data from robotic surgeries offers promising avenues for improving surgical safety, efficiency, and documentation~\cite{chen2023real}. However, building reliable AI systems for such high-stakes environments remains a considerable challenge due to the complex and dynamic nature of surgical scenes~\cite{https://doi.org/10.1002/ags3.12513}.

Video object segmentation models, particularly those based on vision foundation models such as Segment Anything Model 2 (SAM2), have emerged as promising tools for surgical video understanding. These models can propagate object-level information across long video sequences while maintaining spatial and temporal consistency~\cite{gao2023deep}. In particular, SAM2 and several of its medical variants such as MedSAM2 and Sugical SAM2 leverage transformer-based attention mechanisms to perform zero-shot segmentation and tracking across frames, making it highly adaptable to the medical field where labeled data are limited~\cite{geetha2024samsam2exploring}. In the context of surgery, such models could support applications ranging from instrument tracking and anatomy localization to retrospective analysis of procedural steps~\cite{jiaxing2025sam2imagevideosegmentation}.

Tracking performance in video object segmentation heavily depends on the type of user input used to initiate and guide object propagation. Tracking can be performed in three ways: (1) via full segmentation masks, (2) via bounding boxes, and (3) via tracking points~\cite{ravi2024sam}. Among these, segmentation masks offer the most accurate object representation but are time-consuming and expensive to generate, especially in surgical videos where an expert is needed for the annotation~\cite{kulyabin2024segment,liu2024surgical}. Bounding boxes, while more convenient, are often ill-suited for fine-grained surgical tracking due to their coarse spatial localization. Point-based tracking, which involves selecting a few reference points on the object of interest, offers an efficient and low-cost alternative. Several recent studies have suggested that, under ideal conditions, point-based tracking has the potential to reach segmentation-level accuracy, making it an attractive target for real-world surgical applications~\cite{rajivc2025segment}. Despite its practical appeal, the reliability and failure modes of point-based tracking in complex surgical scenarios remain poorly understood. In this work, we tackle this gap by conducting a systematic investigation into the performance and failure modes of point-based tracking in complex surgical environments. To that end, we systematically investigate the failure modes of SAM2 when used for point-based object tracking in surgical videos, focusing on laparoscopic cholecystectomy.

Our results show that point-based tracking is almost always inferior to segmentation mask-based tracking, particularly for anatomical targets. Notably, we observe that point-based tracking fails in distinct ways for the gallbladder, exhibiting issues such as shape misalignment and failure to adapt to anatomical deformation. In contrast, for surgical instruments such as the grasper and L-hook electrocautery, point-based tracking performs considerably better, sometimes approaching the accuracy of segmentation mask-based tracking. Finally, we present a range of qualitative examples highlighting these failure modes, providing practical insights into when and why point-based tracking succeeds or fails in surgical video analysis.

\section{Methodology}
\label{sec:methodology}

\subsection{Model}

Recent advances in VOS models, particularly transformer-based foundation models, have shown strong zero-shot performance in complex domains such as surgical video analysis~\cite{https://doi.org/10.1002/ags3.12513,geetha2024samsam2exploring}. In this work, we leverage SAM2.1 Hiera Large~\cite{ravi2024sam}, a state-of-the-art zero-shot segmentation model that employs a hierarchical transformer architecture for robust spatial and temporal reasoning. SAM2 has shown remarkable performance in medical image segmentation and surgical video understanding~\cite{dosovitskiy2020image,ozbulak2025revisitingevaluationbiasintroduced,vaswani2017attention}, thanks in part to its built-in tracking mechanism that reduces frame-wise segmentation drift and ensures stable object propagation over time. This temporal consistency is particularly important for surgical video analysis, where maintaining coherent tracking of instruments and anatomical structures supports real-time decision-making and situational awareness in the operating room~\cite{10.1001/jamasurg.2019.4917}.

Several adaptations of SAM2 have been proposed for medical applications, including MedSAM2 for static medical images and Surgical-SAM2 for video-based surgical tracking~\cite{liu2024surgical,ma2025medsam2segment3dmedical}. Surgical-SAM2 improves computational efficiency through frame pruning and iterative refinement, making it well suited for offline or sparse-frame video processing~\cite{liu2024surgical}. However, in this study, we intentionally avoid using Surgical-SAM2 to preserve the native 25 frames per second (FPS) video streams without pruning, allowing us to assess the tracking performance of SAM2 in continuous, real-time surgical video streams.

\subsection{Data}
\label{sec:data}

For this study, we use a subset of the CholecSeg8k dataset~\cite{Czempiel2022}, a frame-level annotated surgical video dataset widely used for evaluating segmentation and tracking models in laparoscopic procedures. While CholecSeg8k contains a broader set of annotated videos and object categories, we specifically select ten video segments and three target objects to conduct a focused and detailed analysis of tracking performance. This subset strikes a balance between covering sufficient surgical variability and allowing for a thorough qualitative and quantitative study of failure cases.

We focus our analysis on three key targets encountered during cholecystectomy procedures: the gallbladder, the grasper, and the L-hook electrocautery instrument. These objects represent two distinct tracking challenges, deformable anatomy and rigid surgical tools, providing two diverse scenarios for evaluating the robustness of point-based tracking. We intentionally exclude additional surgical targets, such as the liver, connective tissue, and fat, to maintain a clear experimental scope and to avoid diluting our analysis across too many object categories. This allows us to concentrate on a representative yet manageable set of tracking scenarios that reveal distinct failure modes.

\subsection{Tracking Points}
\label{sec:tracking}

To investigate how the choice and placement of tracking points affect performance, we generate tracking points using three distinct strategies: k-medoids clustering, Shi-Tomasi corner detection, and random sampling.

\begin{itemize}
    \item \textbf{K-medoids}: Points are placed near the geometric centers of the object mask, resembling centroid-based initialization~\cite{kaur2014k}.
    \item \textbf{Shi-Tomasi}: Points are placed close to the edges of the object~\cite{bansal2021efficient}.
    \item \textbf{Random}: Points are sampled uniformly from the object mask, simulating arbitrary point selection without prior knowledge.
\end{itemize}

More details about point selection using k-medoids clustering and Shi-Tomasi corner detection can be found in the work of ~\cite{rajivc2025segment}.

For each strategy, we run experiments using 20 different random seeds, allowing us to assess variability due to different point placements. Additionally, we evaluate the impact of the number of tracking points by varying the quantity across five settings: 1, 2, 3, 5, and 7 points.

\section{Experimental Results}

\begin{table}[t!]
\centering
\scriptsize
\caption{Highest average IoU scores for each object and video segment across different tracking point configurations. For each row, we highlight the best-performing setup in \textbf{bold font} and \underline{underline} the highest score among the point-based tracking configurations. The \textit{Segmentation Mask} column reports the IoU when the full object mask is used for tracking initialization. Results for 1, 2, 3, 5, and 7 points correspond to the best score obtained across all point selection strategies.}

\label{tab:iou_per_video}
\renewcommand{\arraystretch}{1.2}
\begin{tabular}{
>{\centering\arraybackslash}m{2.2cm}
>{\centering\arraybackslash}m{1.3cm}
>{\centering\arraybackslash}m{1.8cm}
>{\centering\arraybackslash}m{1.1cm}
>{\centering\arraybackslash}m{1.1cm}
>{\centering\arraybackslash}m{1.1cm}
>{\centering\arraybackslash}m{1.1cm}
>{\centering\arraybackslash}m{1.1cm}
}
\toprule
\multirow{2}{*}{\shortstack{Tracking\\Target}}  & \multirow{2}{*}{\shortstack{Video\\Segment}}  & \multirow{2}{*}{\shortstack{Segmentation\\Mask}} & \multicolumn{5}{c}{Number of Tracking Points} \\
\cmidrule(lr){4-8}
& & & 1 pt & 2 pts & 3 pts & 5 pts & 7 pts \\
\midrule
\multirow{10}{*}{Gallbladder}
& V01-S4 & \textbf{85.3} & 57.8 & 52.4 & 52.0 & 51.7 & \underline{62.1} \\
& V17-S2 & \textbf{95.3} & 94.3 & 94.0 & 94.4 & \underline{94.9} & \underline{94.9} \\
& V18-S1 & \textbf{88.2} & 67.4 & 66.0 & 60.8 & 67.7 & \underline{75.7} \\
& V20-S1 & \textbf{93.1} & 42.4 & 43.8 & 43.5 & \underline{57.1} & 55.4 \\
& V24-S5 & \textbf{96.1} & 51.3 & 51.5 & 55.4 & 54.1 & \underline{55.8} \\
& V26-S2 & 91.4 & 88.4 & 91.0 & 91.2 & 91.3 & \underline{\textbf{91.7}} \\
& V35-S1 & \textbf{93.7} & 86.9 & 89.7 & 89.4 & 90.0 & \underline{90.2} \\
& V35-S2 & \textbf{95.0} & 92.9 & 93.2 & 93.2 & 93.5 & \underline{94.3} \\
& V35-S3 & \textbf{92.1} & 90.9 & 90.8 & 90.7 & 90.4 & \underline{91.2} \\
& V48-S2 & 91.6 & \underline{94.6} & 94.4 & \underline{94.6} & \underline{94.6} & \textbf{\underline{94.6}} \\
\midrule
\multirow{10}{*}{Grasper}
& V01-S4 & \textbf{81.9} & 63.0 & 65.8 & 65.6 & 68.1 & \underline{68.4} \\
& V17-S2 & \textbf{86.1} & 84.1 & 84.0 & 84.2 & \underline{84.4} & 84.3 \\
& V18-S1 & 94.3 & 94.7 & \textbf{\underline{94.8}} & 94.7 & 94.7 & 94.7 \\
& V20-S1 & \textbf{94.6} & 93.4 & 93.5 & 93.6 & \underline{93.8} & \underline{93.8} \\
& V24-S5 & 90.6 & 90.8 & 90.9 & 91.0 & \textbf{\underline{91.2}} & 91.0 \\
& V26-S2 & \textbf{77.3} & 75.5 & \underline{75.6} & 75.4 & 75.1 & 75.2 \\
& V35-S1 & \textbf{94.3} & 94.2 & \underline{94.3} & \underline{94.3} & \underline{94.3} & \underline{94.3} \\
& V35-S2 & \textbf{93.1} & \underline{92.4} & \underline{92.4} & \underline{92.4} & \underline{92.4} & 92.3 \\
& V35-S3 & \textbf{96.2} & 95.9 & 95.9 & 95.9 & 95.9 & \underline{96.0} \\
& V48-S2 & \textbf{80.1} & 76.4 & 75.7 & \underline{76.8} & 75.9 & 76.6 \\
\midrule
\multirow{3}{*}{\shortstack{L-hook\\Electrocautery}}
& V01-S4 & 90.8 & 92.3 & \underline{\textbf{92.6}} & 89.9 & 89.9 & 90.0 \\
& V18-S1 & \textbf{94.4} & 93.4 & 93.4 & \underline{93.5} & \underline{93.5} & \underline{93.5} \\
& V20-S1 & \textbf{91.7} & 90.9 & 91.0 & 90.9 & 91.0 & \underline{91.2} \\

\bottomrule
\end{tabular}
\vspace{-1em}
\end{table}

Using the setup described in Section~\ref{sec:methodology}, we conduct experiments to evaluate the tracking performance of SAM2 across different objects, videos, and tracking point configurations. Our goal is to compare point-based tracking against segmentation mask-based tracking and analyze how the number and placement of points influence performance.

In Table~\ref{tab:iou_per_video}, we present the highest average IoU scores obtained for each object and video segment across all point selection strategies. For each video, we report the best result achieved with 1, 2, 3, 5, or 7 points, allowing us to isolate the upper performance limit of point-based tracking irrespective of initialization method. Our observations from this table are as follows:

\begin{itemize}
    \item For the gallbladder, segmentation mask-based tracking consistently outperforms point-based tracking across all videos, highlighting the challenges of tracking anatomical structures with tracking points.
    
    \item For the grasper and L-hook electrocautery, point-based tracking achieves competitive performance, with several videos showing minimal difference between points and segmentation. This suggests that surgical tools are easier to track with tracking points.
    
    \item The gap between segmentation and point-based tracking is larger for anatomical targets than for surgical tools.
    
    \item Increasing the number of points generally improves tracking performance, but does not fully close the gap with segmentation masks for anatomical targets. For surgical tools, even a small number of points can yield strong results.
\end{itemize}

\begin{figure}[htbp!]
\centering
\begin{subfigure}{\textwidth}
\includegraphics[width=0.32\textwidth]{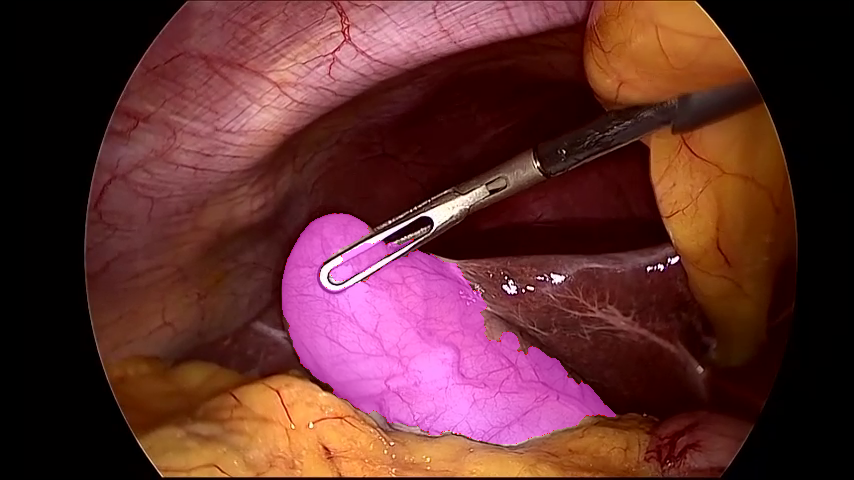}
\includegraphics[width=0.32\textwidth]{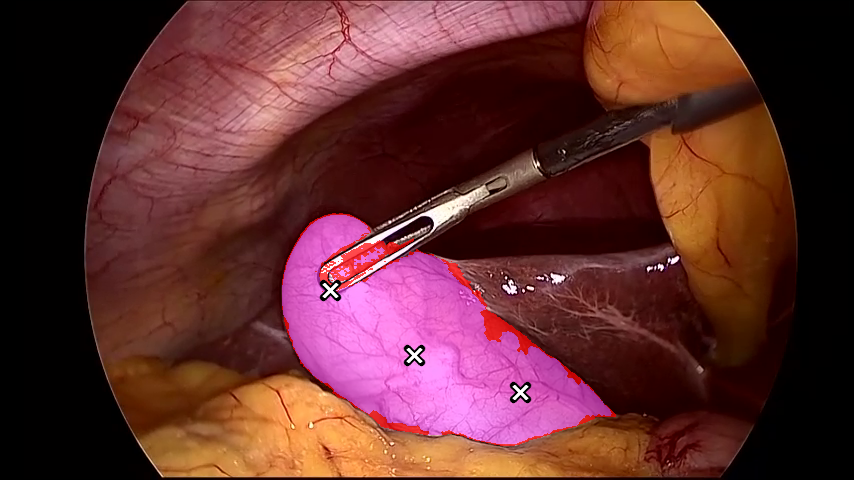}
\includegraphics[width=0.32\textwidth]{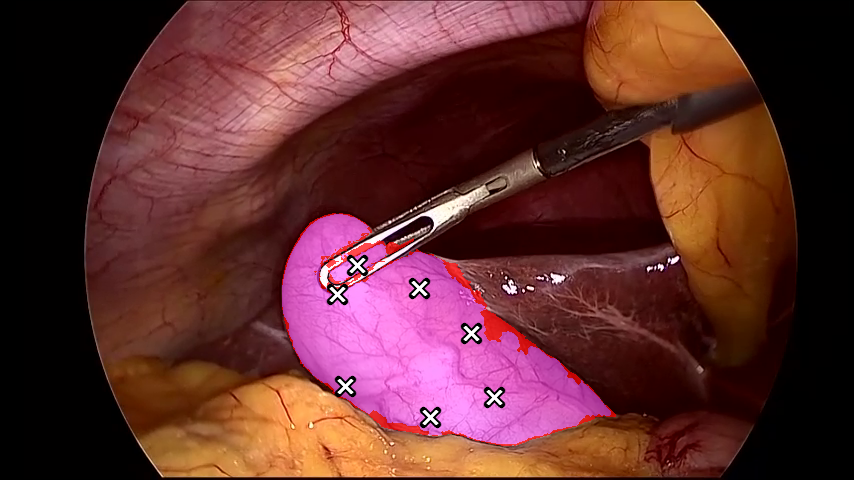}
\\
\includegraphics[width=0.32\textwidth]{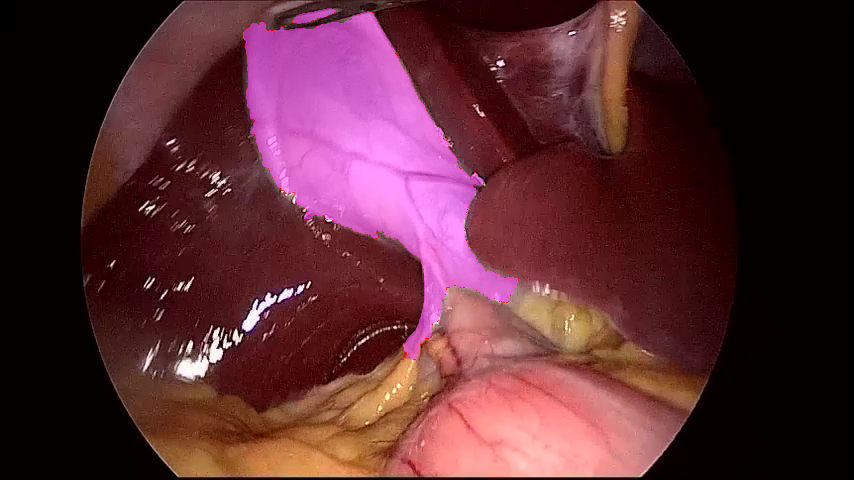}
\includegraphics[width=0.32\textwidth]{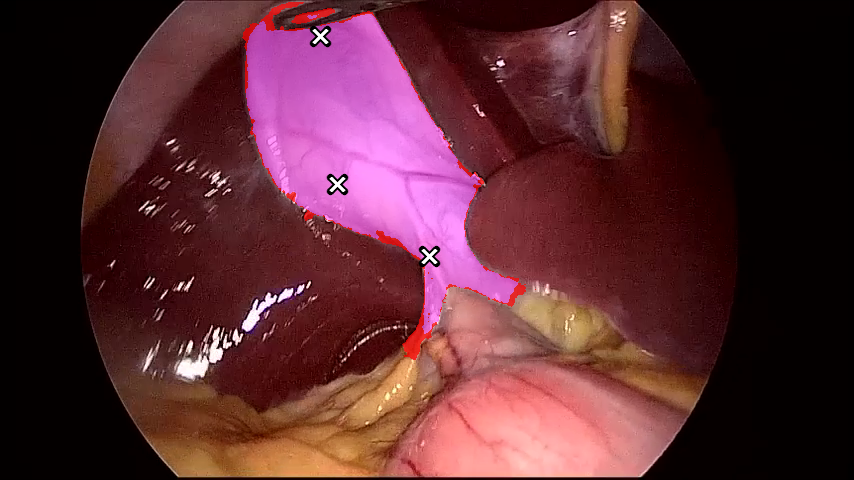}
\includegraphics[width=0.32\textwidth]{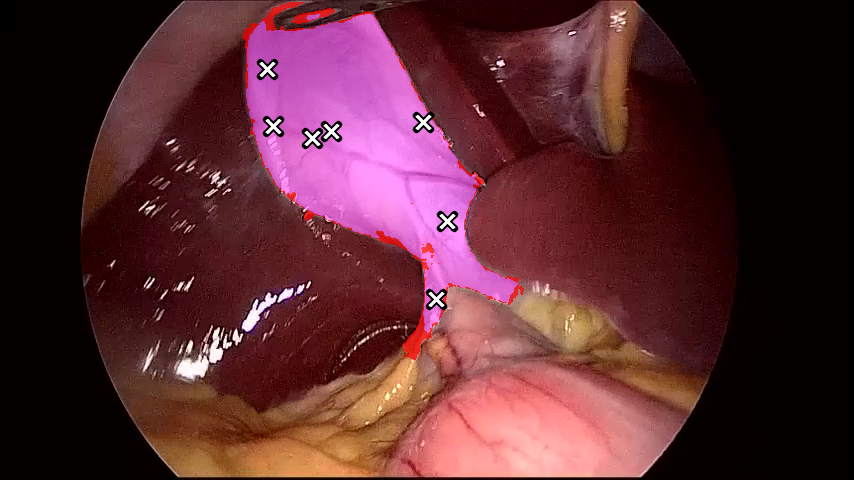}
\\
\includegraphics[width=0.32\textwidth]{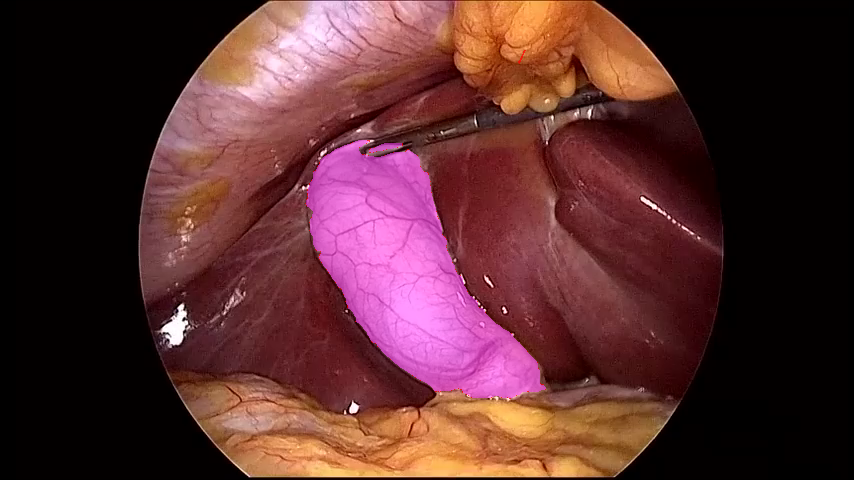}
\includegraphics[width=0.32\textwidth]{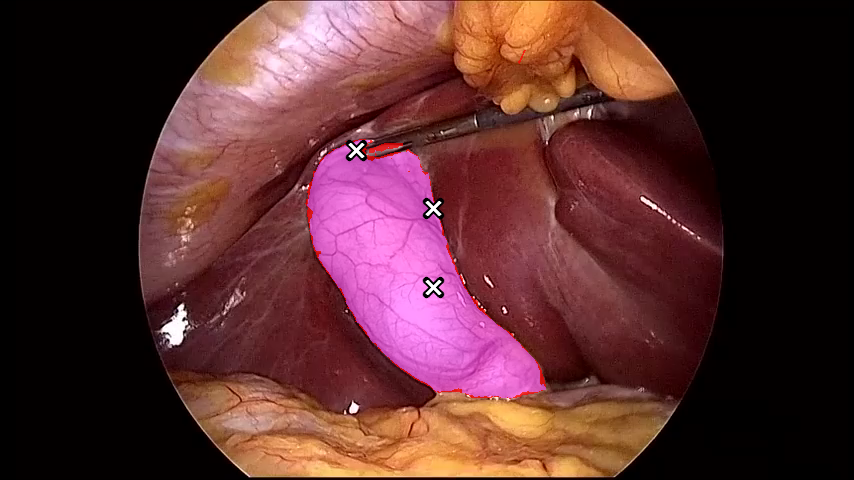}
\includegraphics[width=0.32\textwidth]{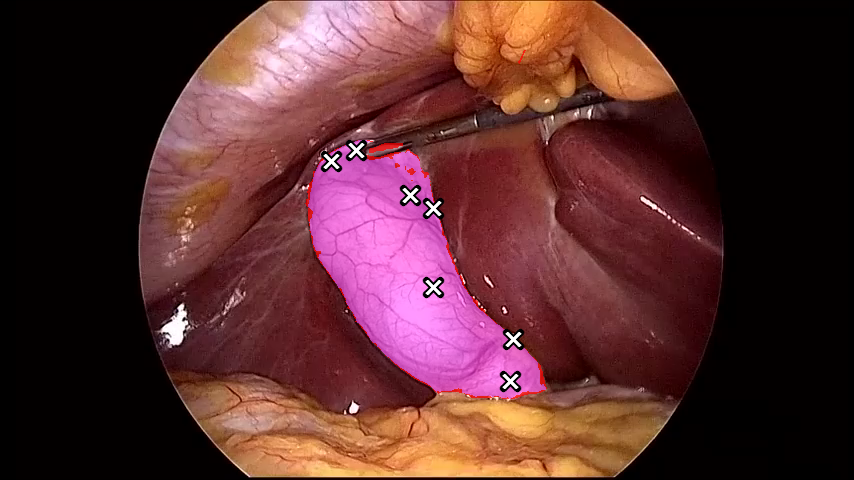}
\caption{Point-based tracking is comparable to mask-based tracking.}
\end{subfigure}
\\
\begin{subfigure}{\textwidth}
\includegraphics[width=0.32\textwidth]{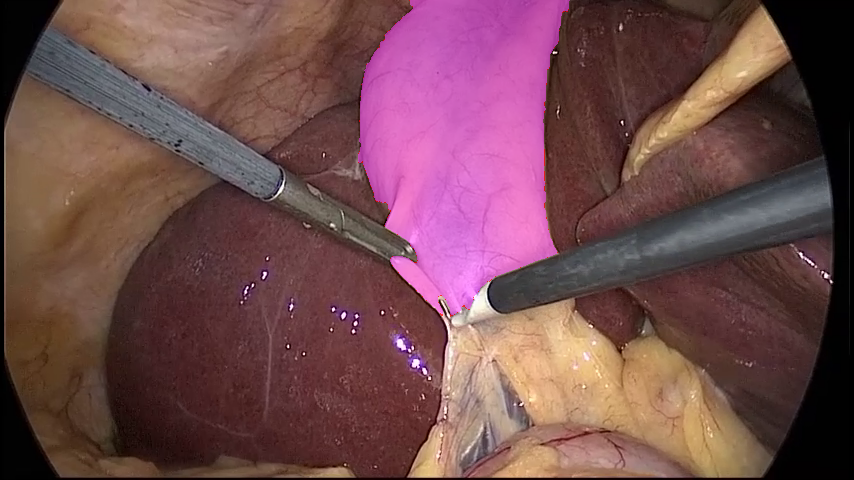}
\includegraphics[width=0.32\textwidth]{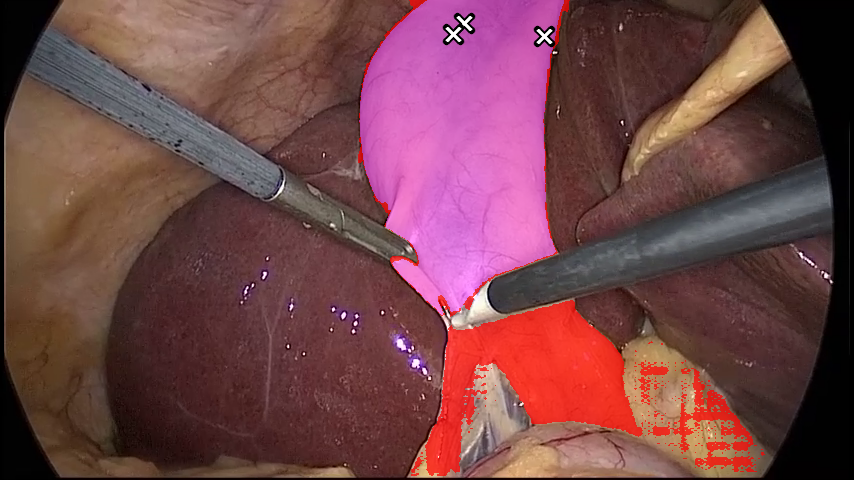}
\includegraphics[width=0.32\textwidth]{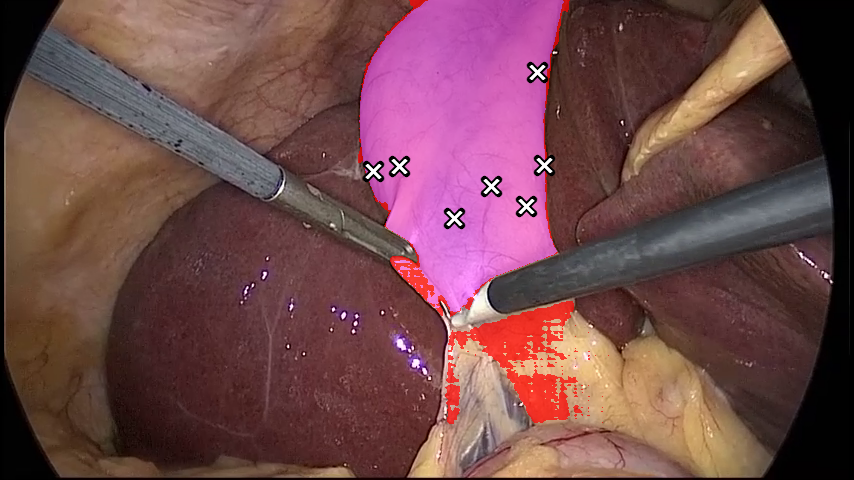}
\\
\includegraphics[width=0.32\textwidth]{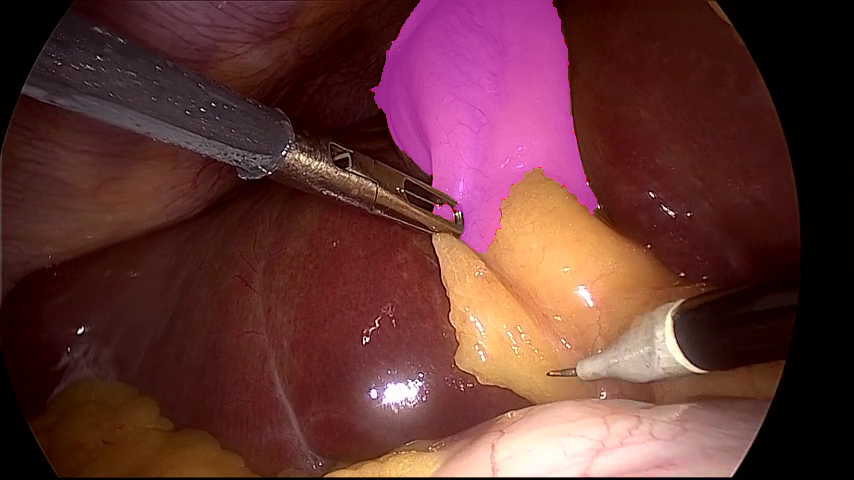}
\includegraphics[width=0.32\textwidth]{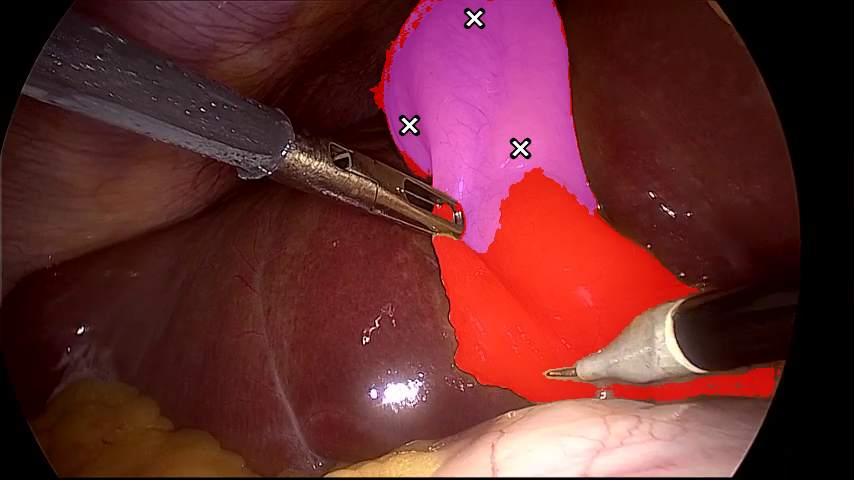}
\includegraphics[width=0.32\textwidth]{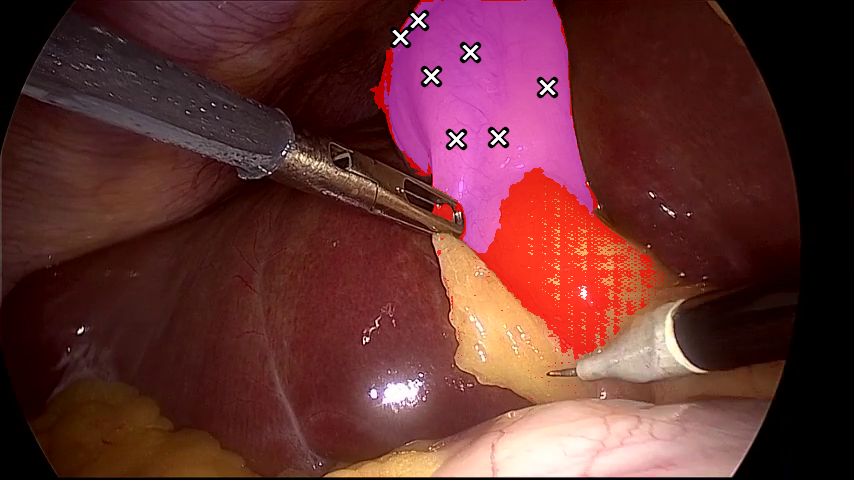}
\\
\includegraphics[width=0.32\textwidth]{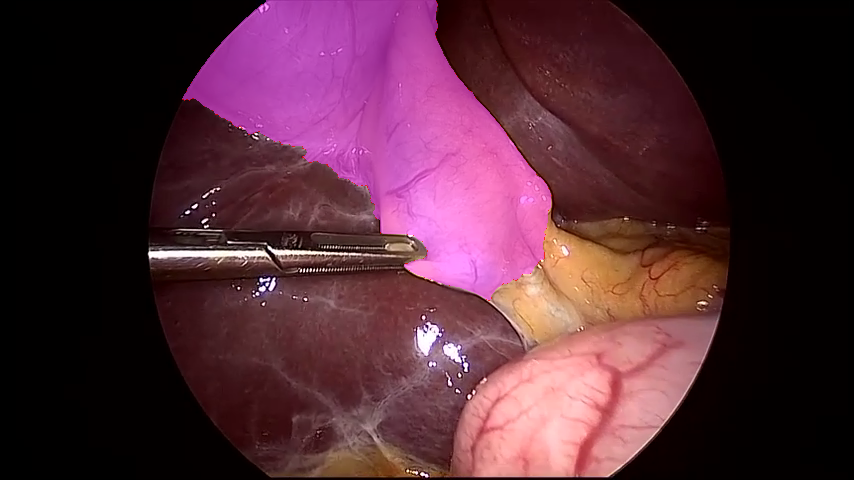}
\includegraphics[width=0.32\textwidth]{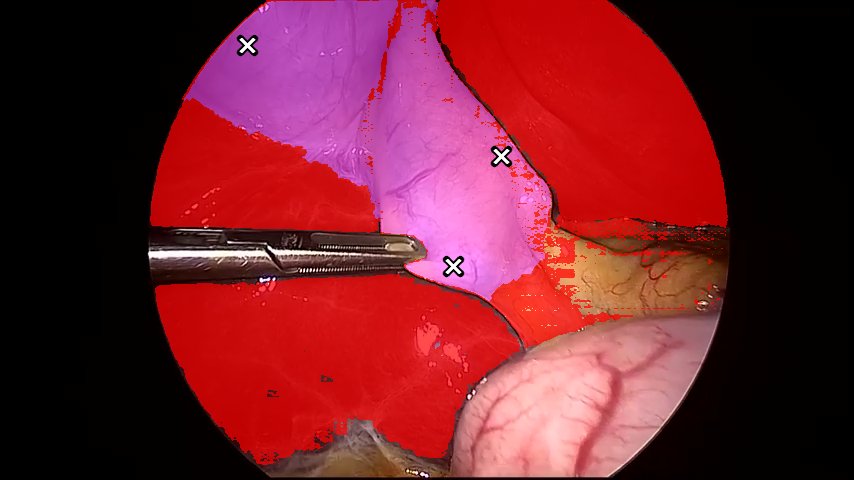}
\includegraphics[width=0.32\textwidth]{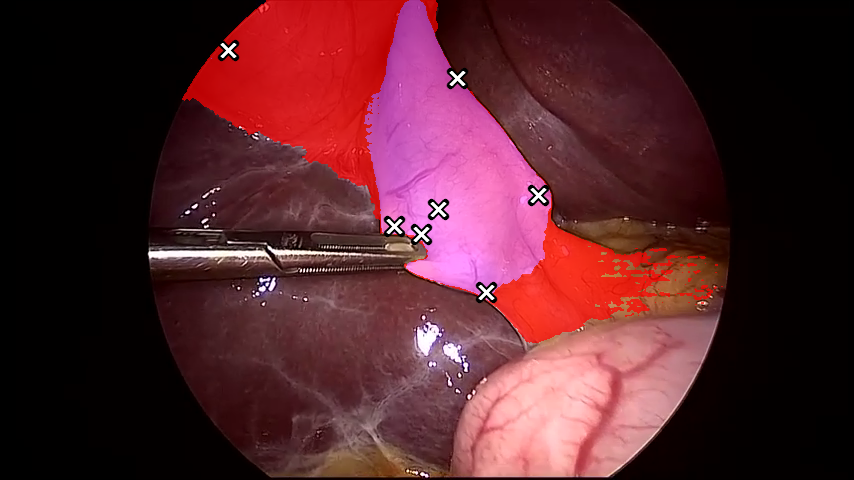}
\caption{Mask-based tracking is substantially better than point-based tracking.}
\end{subfigure}
\caption{Qualitative examples illustrating two distinct tracking outcomes (a) and (b) for the gallbladder. Images in the first column show the ground truth segmentation masks; the second column shows the highest IoU case with 3 tracking points, and the third column shows the highest IoU case with 7 tracking points. Gallbladder regions correctly predicted are highlighted in pink, and incorrect predictions are highlighted in red. Tracking points are shown as white X marks with black outlines.}
\label{fig:gallbladder_examples}
\end{figure}

\begin{figure}[t!]
\centering
\begin{subfigure}{\textwidth}
\centering
\begin{subfigure}{0.48\textwidth}
\includegraphics[width=0.49\textwidth]{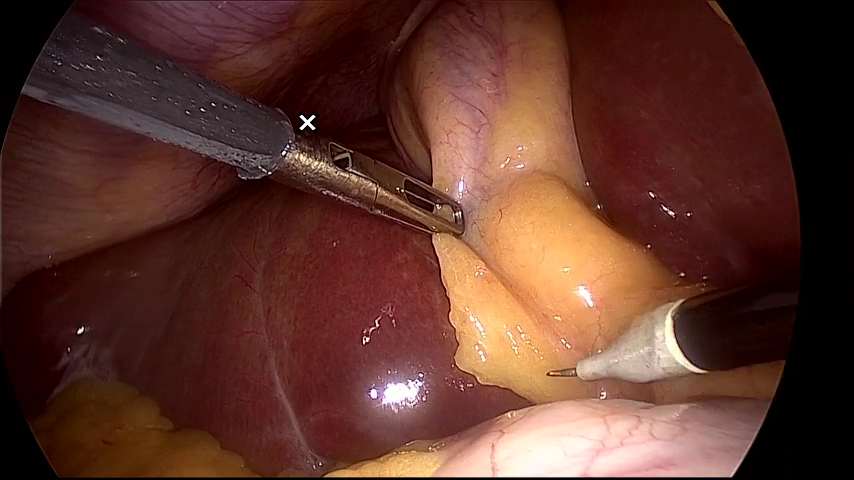}
\includegraphics[width=0.49\textwidth]{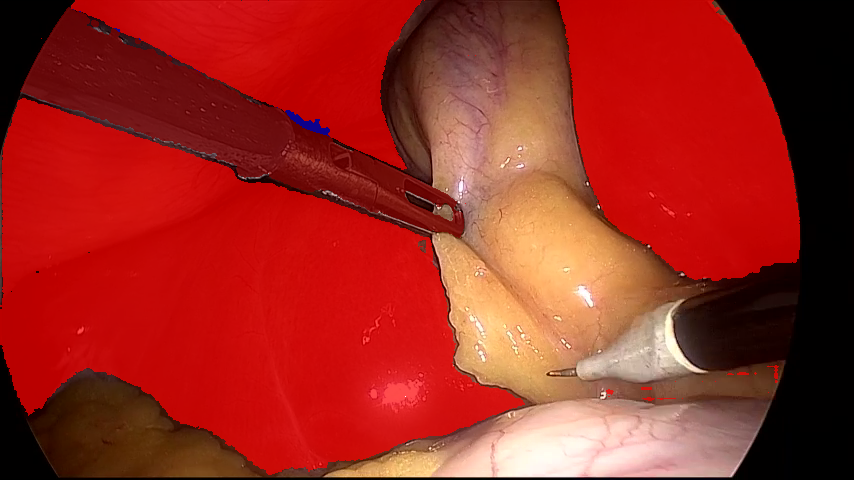}
\\
\includegraphics[width=0.49\textwidth]{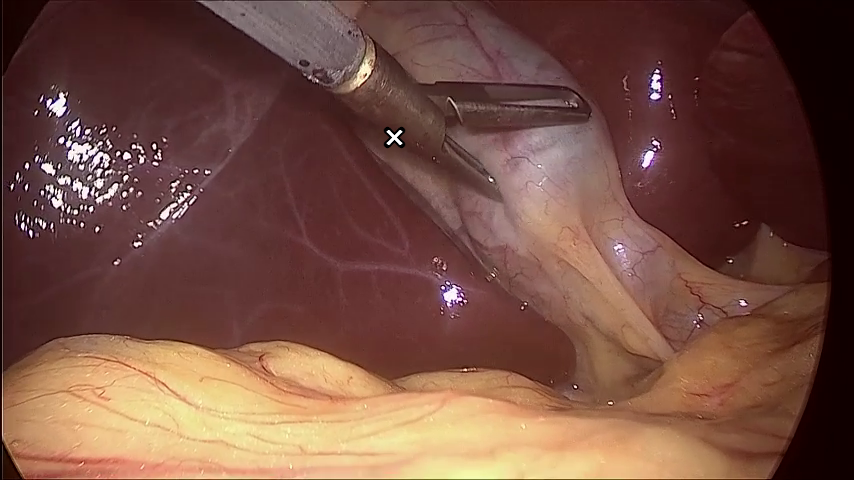}
\includegraphics[width=0.49\textwidth]{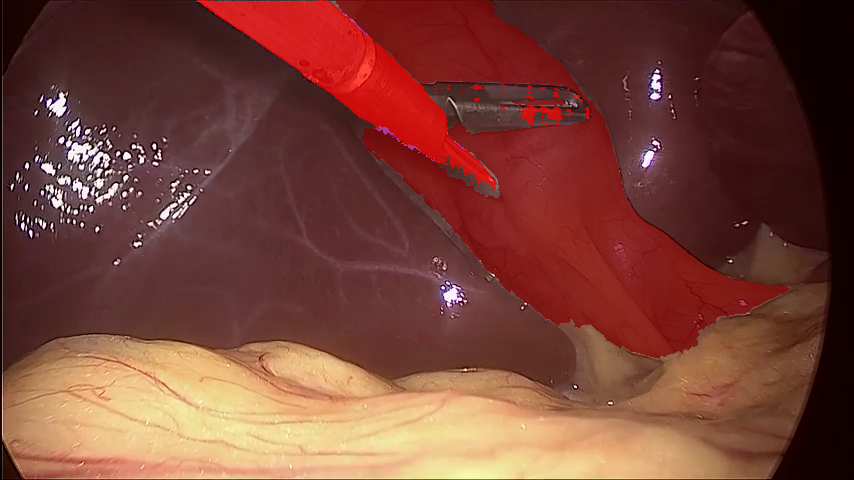}
\end{subfigure} 
\hspace{1mm}
\begin{subfigure}{0.48\textwidth}
\includegraphics[width=0.49\textwidth]{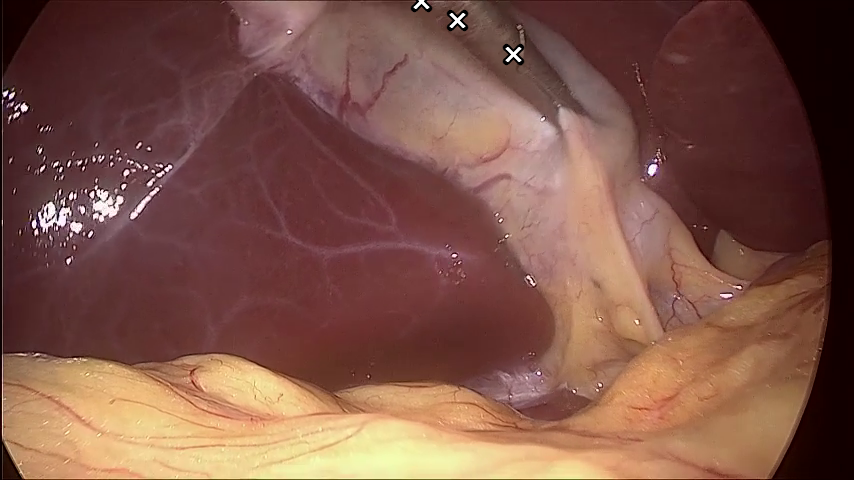}
\includegraphics[width=0.49\textwidth]{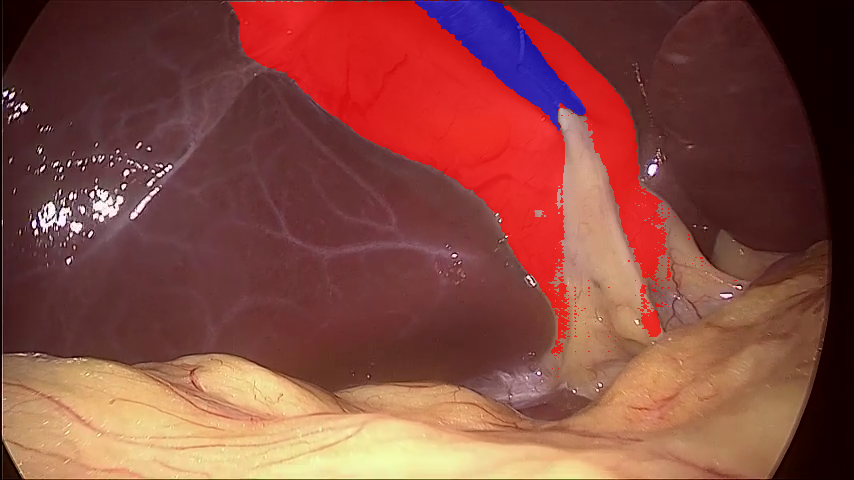}
\\
\includegraphics[width=0.49\textwidth]{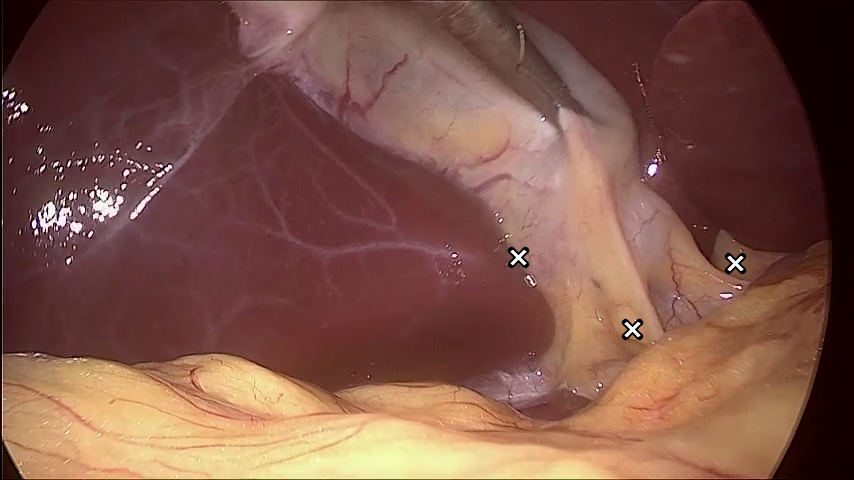}
\includegraphics[width=0.49\textwidth]{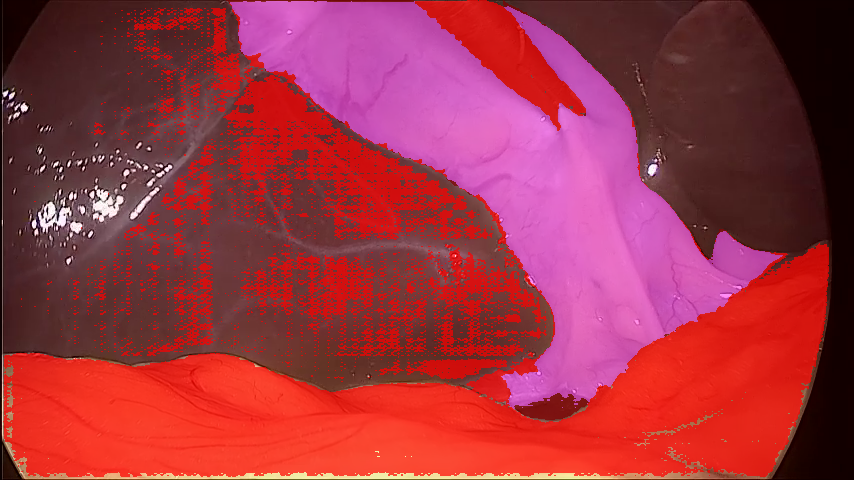}
\end{subfigure} 
\caption{Tracking points closer to the edge of target objects}
\end{subfigure} 
\begin{subfigure}{\textwidth}
\centering
\begin{subfigure}{0.48\textwidth}
\includegraphics[width=0.49\textwidth]{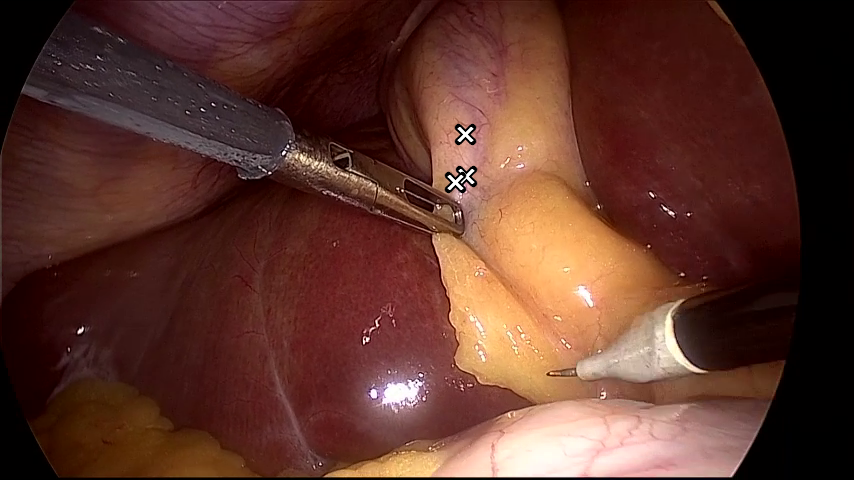}
\includegraphics[width=0.49\textwidth]{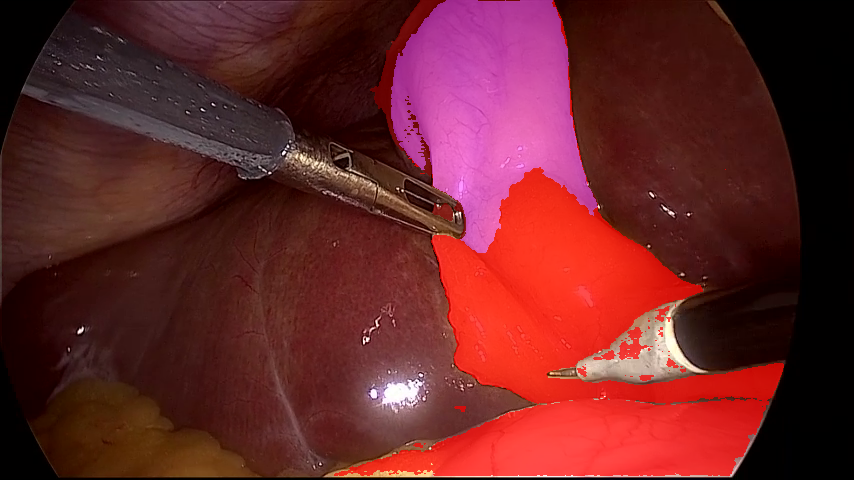}
\\
\includegraphics[width=0.49\textwidth]{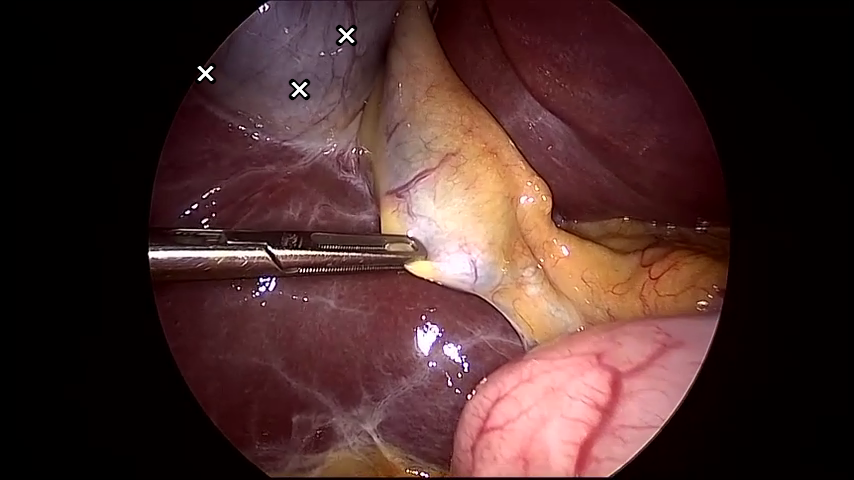}
\includegraphics[width=0.49\textwidth]{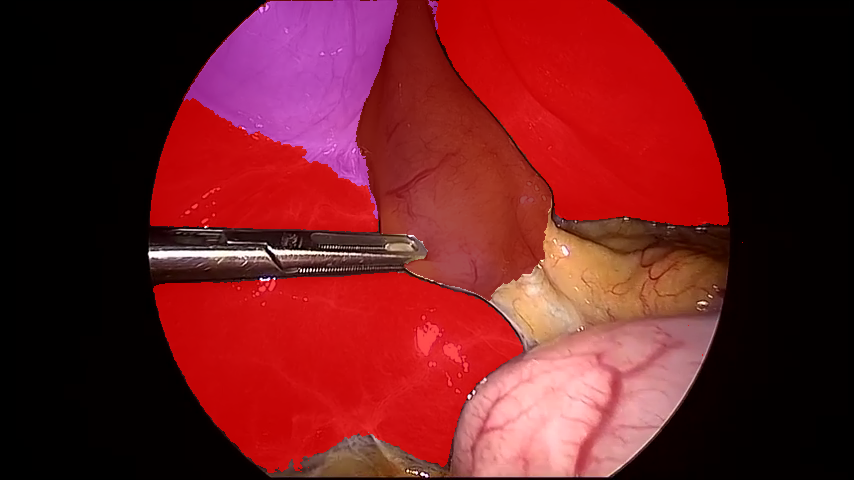}
\end{subfigure}
\hspace{1mm}
\begin{subfigure}{0.48\textwidth}
\includegraphics[width=0.49\textwidth]{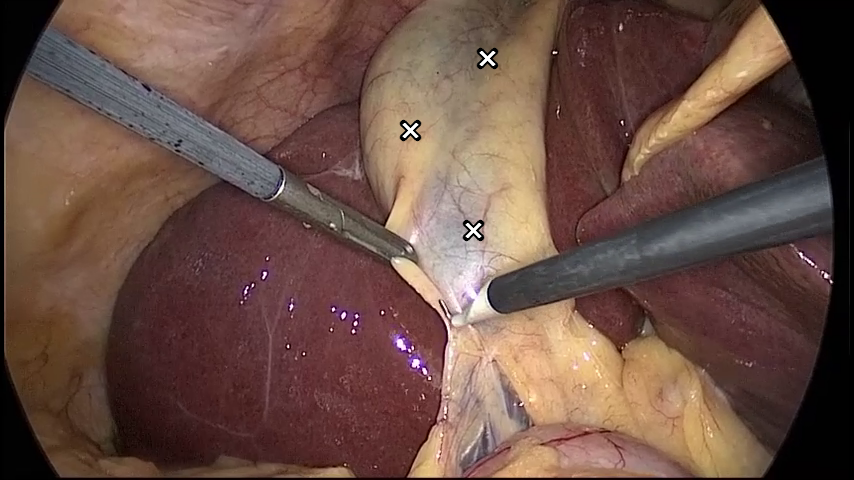}
\includegraphics[width=0.49\textwidth]{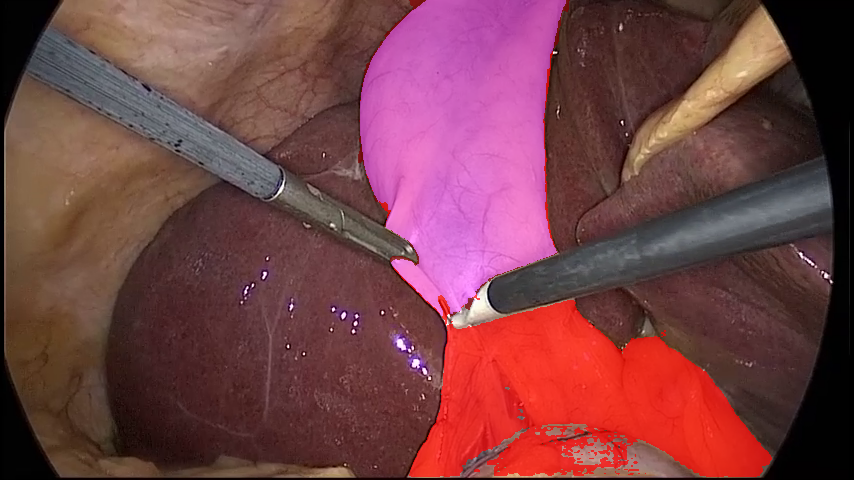}
\\
\includegraphics[width=0.49\textwidth]{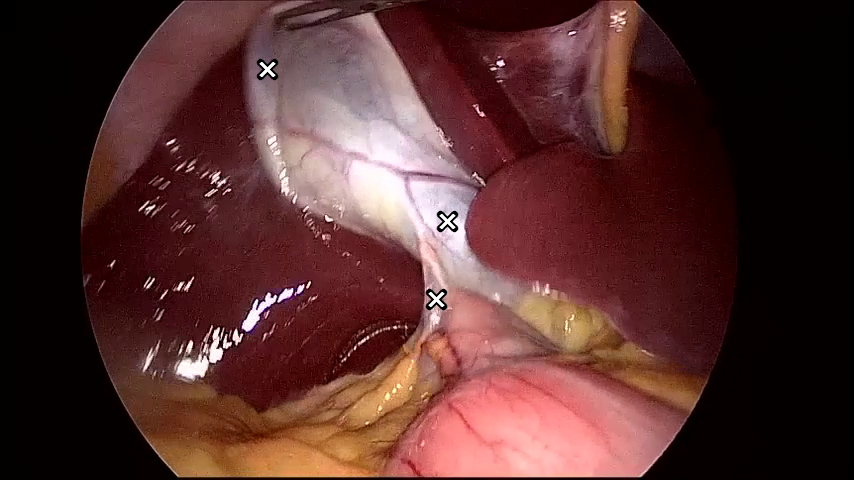}
\includegraphics[width=0.49\textwidth]{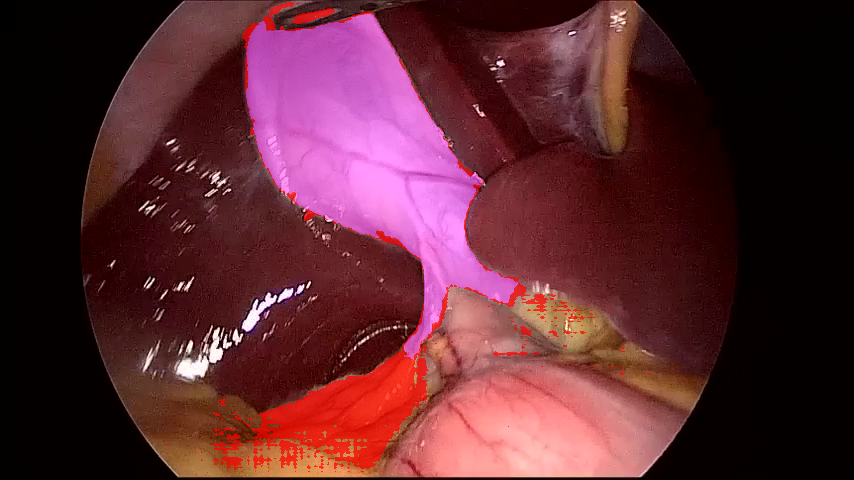}
\end{subfigure}
\caption{Tissue similarity}
\end{subfigure}

\vspace{1mm}

\begin{subfigure}{\textwidth}
\centering
\begin{subfigure}{0.48\textwidth}
\includegraphics[width=0.49\textwidth]{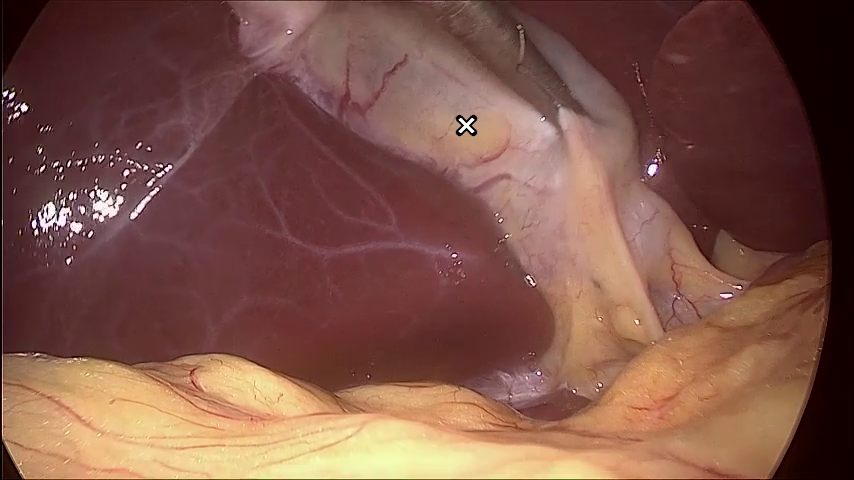}
\includegraphics[width=0.49\textwidth]{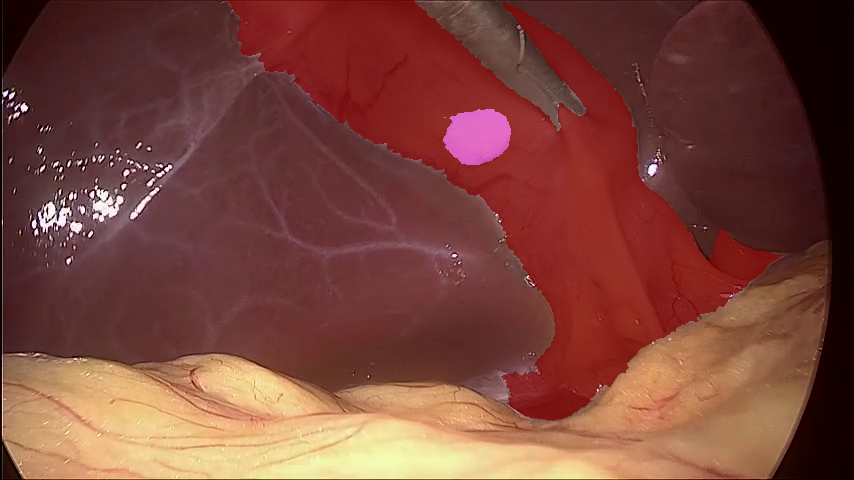}
\\
\includegraphics[width=0.49\textwidth]{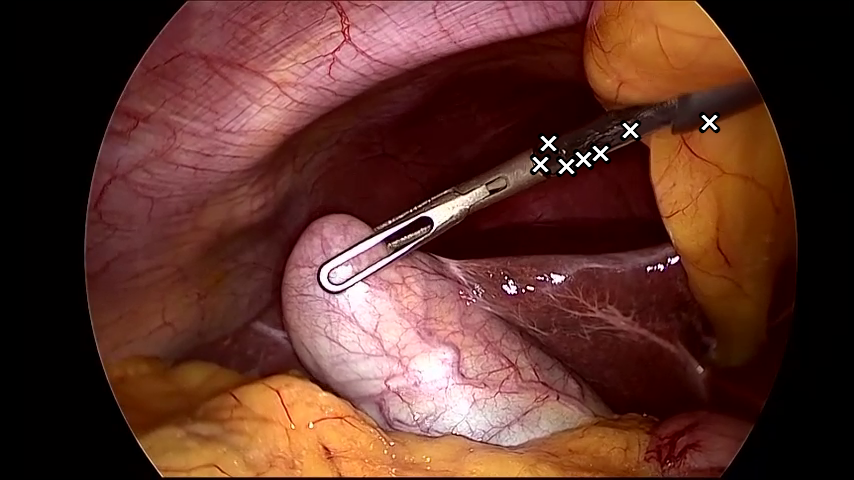}
\includegraphics[width=0.49\textwidth]{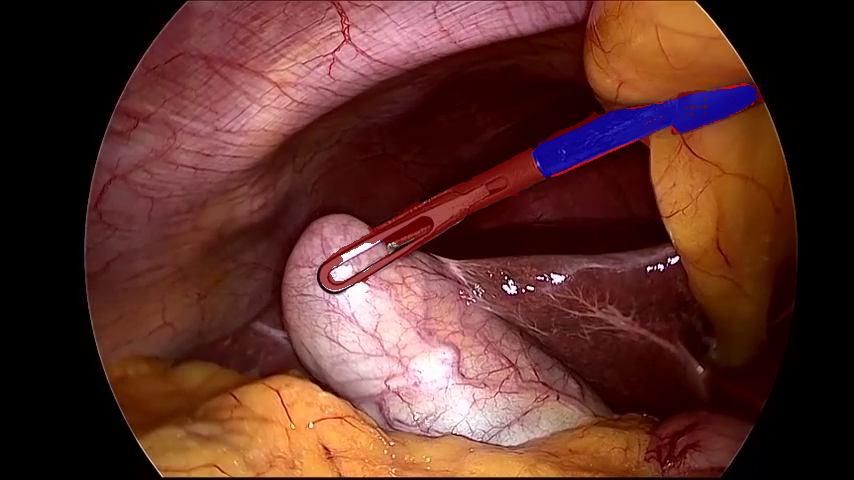}
\end{subfigure}
\hspace{1mm}
\begin{subfigure}{0.48\textwidth}
\includegraphics[width=0.49\textwidth]{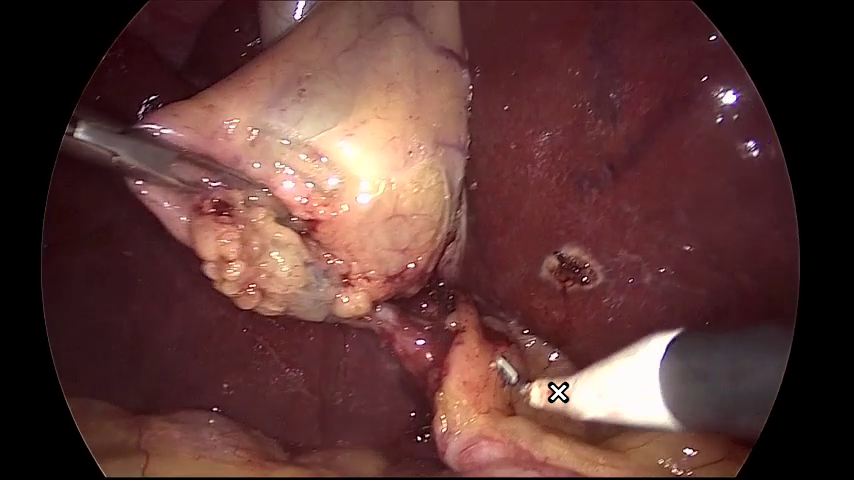}
\includegraphics[width=0.49\textwidth]{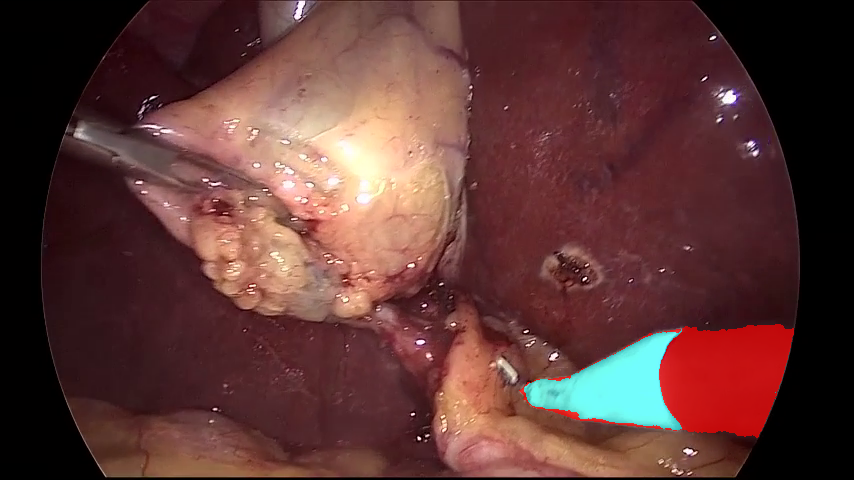}
\\
\includegraphics[width=0.49\textwidth]{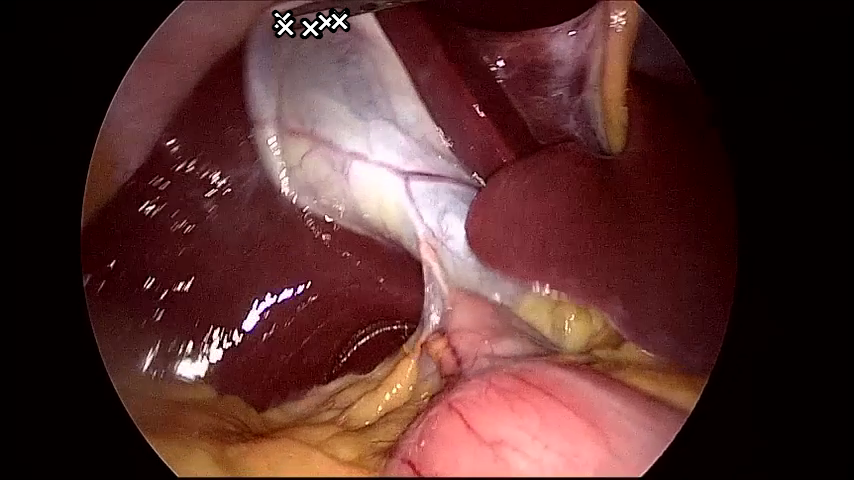}
\includegraphics[width=0.49\textwidth]{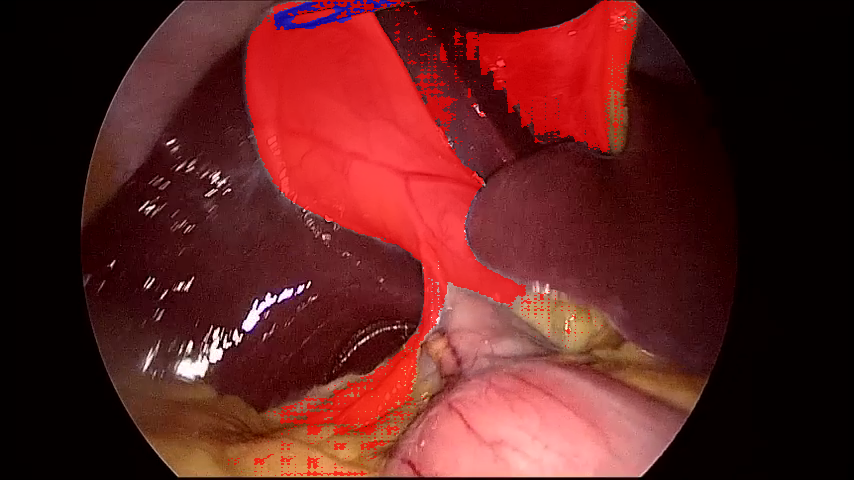}
\end{subfigure}
\caption{Extraordinary cases}
\end{subfigure}

\caption{Qualitative examples illustrating three distinct failure modes in point-based tracking for the gallbladder. For each image pair, the left image represents the input with tracking points and thee right one the segmentation output. (a) Failures due to tracking points placed near the object edges, causing the model to lose the target boundary. (b) Failures caused by tissue similarity, where surrounding structures confuse the model and lead to tracking drift. (c) Extraordinary cases that require case-specific investigation, such as partial object visibility or ambiguous visual cues. Incorrect predictions are highlighted in red and correctly predicted regions are highlighted using different colors for three objects (pink, blue, and cyan).}
\label{fig:failure_examples}
\end{figure}

We now perform a deeper investigation of the gallbladder tracking results by categorizing the cases into two groups: (1) cases where segmentation mask tracking and the best point-based tracking achieve comparable IoU scores, and (2) cases where segmentation mask tracking performs substantially better than point-based tracking.

These comparisons are illustrated in Figure~\ref{fig:gallbladder_examples}. As seen in these examples, when the surrounding tissues, such as connective tissue or fat, share similar appearance and texture with the gallbladder, the model often fails to maintain correct point tracking. The tracking points struggle to capture the deformable and indistinct boundaries of the organ, leading to the selection of adjacent regions into the segmentation.

On the other hand, in simpler scenes where the gallbladder has clear boundaries and is well-isolated from surrounding structures, point-based tracking performs competitively. In these cases, even a small number of points can successfully follow the object across frames, indicating that visual separation plays a critical role in tracking success.

Overall, these findings highlight a key failure mode of point-based tracking for anatomical structures: the model struggles when object boundaries are ambiguous or when object appearance blends with nearby tissues. Mask-based tracking, which has full spatial context, is less susceptible to these issues and remains consistently robust in such scenarios.

Upon qualitative inspection, we identify several recurring failure modes in point-based tracking. First, we observe that when tracking points are selected near the edges of the target object, the model often becomes confused, causing the tracked region to drift outside the object boundaries. This issue is observed not only for anatomical structures but also for surgical tools, where selecting points on tool edges sometimes leads to tracking leakage into the background or adjacent objects. Second, as discussed earlier, tissue similarity plays a major role in failure cases: when adjacent tissues closely resemble the tracked object in color or texture, the model struggles to distinguish between them, leading to tracking errors.

In addition to these common patterns, we identify several case-specific failures that require individual investigation. Representative examples of these cases are shown in Figure~\ref{fig:failure_examples}. For instance, when tracking the L-hook electrocautery or the grasper, if the tracking point is placed only on the tip and not on the handle, or vice-versa, the model tends to follow just the tip (or the handle) due to its strong color contrast, ignoring the rest of the tool. Similarly, when tracking with a single point, we frequently observe that the model only tracks a small localized region, failing to capture the full extent of the object. Another common failure occurs when objects are tracked at the edge of the surgical field of view; in these cases, the model exhibits unstable behavior, sometimes losing the object entirely as it partially exits the frame.

\subsection{Practical Recommendations for Point-Based Tracking}

Based on our experimental findings, we provide the following practical suggestions for using video object segmentation models such as SAM2 in surgical videos when relying on tracking points:

\begin{enumerate}
    \item \textbf{For anatomical structures:} Place several tracking points along the edges of the object rather than at the center. This reduces the likelihood of drift caused by boundary ambiguity and tissue similarity.
    
    \item \textbf{For surgical instruments:} Place several points near the center of the instrument, as tools often appear very thin in the field of view. Additionally, distribute tracking points across visually distinct parts of the tool. For example, placing points on both the white/gray tip and the black handle of the grasper helps the model recognize that these parts belong to the same object.
    
    \item \textbf{When initializing tracking near the edge of the field of view:} Exercise caution, as partial visibility and abrupt scene changes at the frame boundary can lead to tracking failures. Ensuring that points are placed on stable, clearly visible regions mitigates these risks.
\end{enumerate}

\section{Conclusion}

In this work, we investigated the failure modes of point-based tracking in surgical video segmentation using SAM2. Through systematic experiments on cholecystectomy videos, we demonstrated that while point-based tracking can achieve competitive performance for surgical tools, it struggles with anatomical structures due to boundary ambiguity and tissue similarity. Our qualitative analysis further revealed that point placement plays a critical role in tracking success.

Future work should explore the use of \textit{negative points}, which indicate areas that should not be tracked. These points could provide valuable context, helping the model avoid drift and improving tracking robustness in anatomically complex scenes. Due to the limited scope of this study, we did not investigate several important anatomical structures such as connective tissue and the liver, which frequently interact with surgical targets and contribute to visual ambiguity. Future research should expand the range of tracked objects to include these challenging tissues, enabling a more comprehensive understanding of model failure modes across diverse anatomical contexts.

\bibliographystyle{splncs04}
\bibliography{COLAS-22}

\end{document}